\documentclass[a4paper,UKenglish,cleveref, autoref, thm-restate]{lipics-v2021}
\usepackage{booktabs}
\usepackage{xspace}
\usepackage{mathtools}
\usepackage[final]{pdfpages}

\usepackage[ruled,vlined,linesnumbered]{algorithm2e} %algo2e
\SetAlgoSkip{}

\newcommand{\oBCM}{\textsc{One-Sided Bipartite Crossing Minimisation}\xspace}

\newcommand{\nY}{n}

\title{Evolutionary Algorithms for One-Sided Bipartite Crossing Minimisation} %TODO Please add

\titlerunning{Evolutionary Algorithms for OBCM} %TODO optional, please use if title is longer than one line

\author{Jakob Baumann}{University of Passau, Germany}{}{https://orcid.org/
	0000-0002-2594-3828}{} 

\author{Ignaz Rutter}{University of Passau, Germany}{}{https://orcid.org/
	0000-0002-3794-4406}{} 

\author{Dirk Sudholt}{University of Passau, Germany}{}{https://orcid.org/
	0000-0001-6020-1646}{} 

\authorrunning{J. Baumann, I. Rutter, and D. Sudholt}

\Copyright{Jakob Baumann, Ignaz Rutter, and Dirk Sudholt}

\ccsdesc[500]{Theory of computation~Design and analysis of algorithms}
\keywords{Mutation Operator, Layered Graphs, Crossing Minimisation} %TODO mandatory; please add comma-separated list of keywords

\category{Poster Abstract}

\nolinenumbers %uncomment to disable line numbering

\EventEditors{Stefan Felsner and Karsten Klein}
\EventNoEds{2}
\EventLongTitle{32nd International Symposium on Graph Drawing and Network Visualization (GD 2024)}
\EventShortTitle{GD 2024}
\EventAcronym{GD}
\EventYear{2024}
\EventDate{September 18--20, 2024}
\EventLocation{Vienna, Austria}
\EventLogo{}
\SeriesVolume{320}
\ArticleNo{48}
\begin{document}
	
	\maketitle
	
	\begin{abstract}
		Evolutionary algorithms (EAs) are universal solvers inspired by principles of natural evolution. In many applications, EAs produce astonishingly good solutions. %As they are able to deal with complex optimisation problems, they show great promise for hard problems encountered in the field of graph drawing.
		To complement recent theoretical advances in the analysis of EAs on graph drawing~\cite{Baumann2024}, we contribute a fundamental empirical study.
		
		We consider the so-called \textsc{One-Sided Bipartite Crossing Minimisation (OBCM)}: given two layers of a bipartite graph and a fixed horizontal order of vertices on the first layer, the task is to order the vertices on the second layer to minimise the number of edge crossings. 
		We empirically analyse the performance of simple EAs for OBCM and compare different mutation operators on the underlying permutation ordering problem: exchanging two elements (\textit{exchange}), swapping adjacent elements (\textit{swap}) and jumping an element to a new position (\textit{jump}). 
		EAs using jumps easily outperform all deterministic algorithms in terms of solution quality after a reasonable number of generations. We also design variations of the best-performing EAs to reduce the execution time for each generation. The improved EAs can obtain the same solution quality as before and run up to 100 times faster.
	\end{abstract}
	
	% \section{Introduction}
	\section{Empirical Performance Comparison}
	We study the mutations swap, exchange, and jump on a simplistic $(1+1)$-type EA~\cite{Oliveto2007Time}, on \oBCM~\cite{LPPNPcomplete}. The $(1+1)$-EAs (Swap-EA, Exchange-EA, Jump-EA) start with a random permutation and apply the corresponding operator $k$ times, following a Poisson distribution with $\lambda=1$. We also consider the randomised local search (RLS), where we set $k = 1$ constant. 
	We compare the EAs to four state-of-the-art algorithms: The Barycenter and Median algorithms~\cite{LPPNPcomplete}, Nagamochi's algorithm~\cite{nagamochi2005improved}, and a heuristic known as Sifting~\cite{matuszewski1999sifting}. Nagamochi's algorithm gives the best theoretical approximation ratio, but its performance was never empirically evaluated; a gap we aim to close with this work. We note that there are other well-performing algorithms, for which evaluations are readily available. We believe that the chosen subset is sufficient for this comparison.

	% \section{Empirical Performance Comparison}
	We performed tests on three different instances, similar to~\cite{demestrescu2001breaking}. %We generated random instances of two types, and we downloaded matrices from the matrix market\footnote{https://math.nist.gov/MatrixMarket/index.html} and converted them to bipartite graphs (edge $i,j$ exists iff $a_{i,j} \neq 0$ in the matrix). 
	Due to space limitations, and as there are no significant differences, we present only the results for \emph{random instances}, with $n=100$ vertices on both layers, where we added each edge with a fixed probability $p$. Note that we also considered differently sized layers and increasing density $p$; the behaviour of RLS/EAs was basically the same, while the other algorithms were slightly affected, most of which was also covered in~\cite{demestrescu2001breaking}. We computed the optimum solution using an ILP~\cite{junger1997Performance}, which can solve instances of size up to $n\approx 190$. EAs and RLS perform a preprocessing step of computing the cross table~\cite{demestrescu2001breaking}, which takes $\Theta(nm)$ steps.
	% They reported they could solve medium-sized instances with a bottom layer of up to 60 vertices. On today's computers this bound extends to about 190 vertices, after which the ILP runs out of memory, although the running time is also rapidly increasing at that point.
	\begin{figure}[t]
		\centering%
		\subcaptionbox{preprocessing is needed for Sifting and EAs/RLS}[0.5\textwidth]{
			\includegraphics[width=0.5\textwidth]{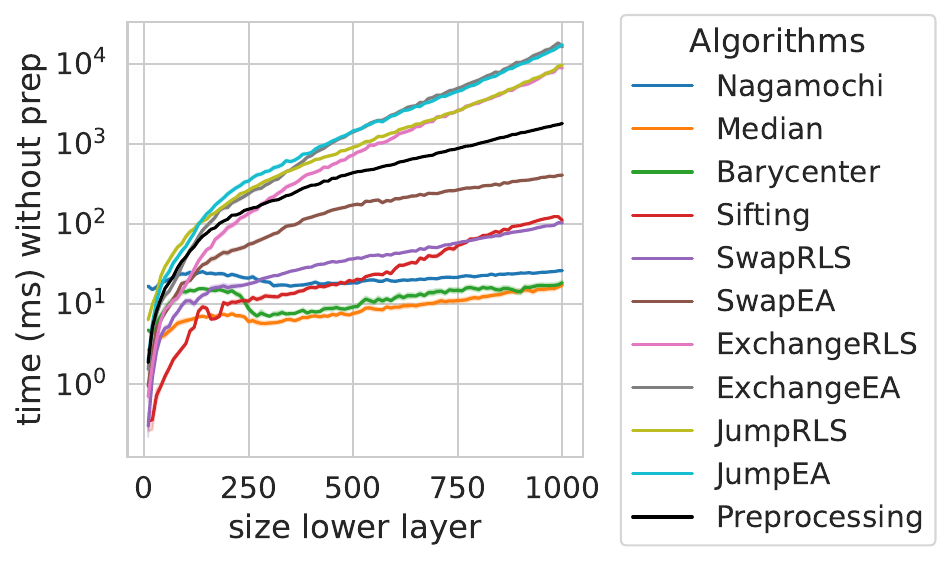}}\medskip\\
		\subcaptionbox{classical algorithms and simple EAs}[0.44\textwidth]{
			\includegraphics[width=0.44\textwidth]{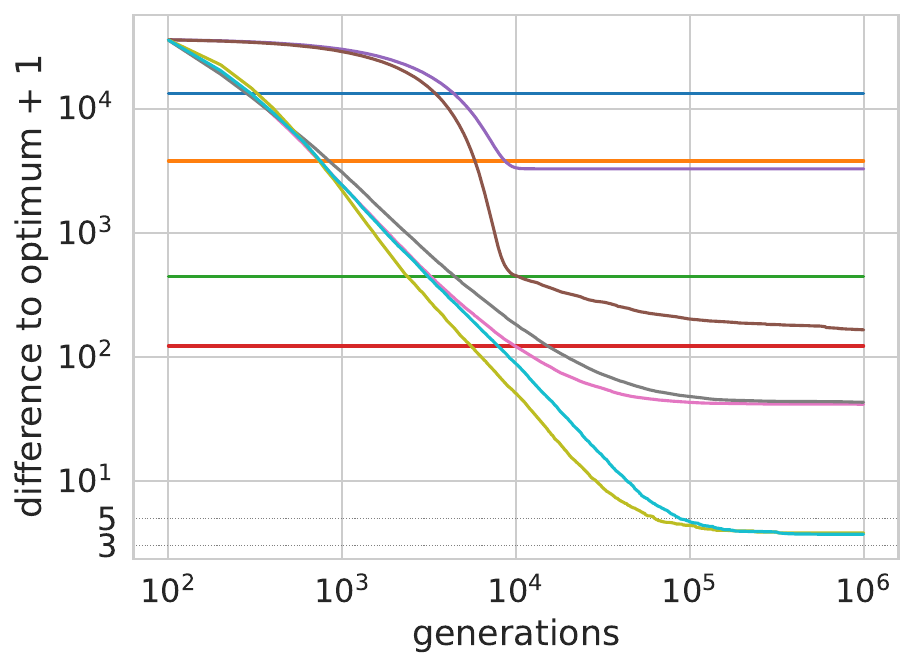}}
		\hfill
		\subcaptionbox{jump variants}[0.44\textwidth]{
			\includegraphics[width=0.44\textwidth]{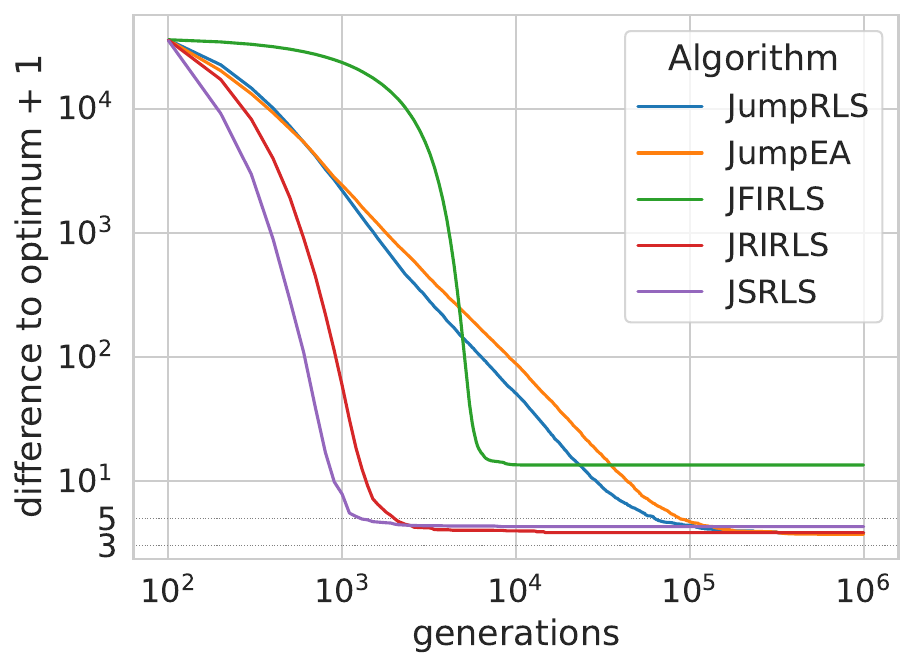}}
		\caption{(a) Wall-clock times averaged over the same 100 random instances with increasing $n$. Evolutionary algorithms were stopped when no improvement was found throughout $\nY^{1.5}$ subsequent generations. The costs for initialising the crossing matrix for the EAs and the Sifting algorithm are subtracted and shown separately. (b) \& (c) Difference between the final evolved crossing number (for EAs) or the returned crossing number (for deterministic algorithms) and the optimal crossing number plotted over generations for classical algorithms and evolutionary algorithms. The plots show averages taken over a suite of instances.}
		\label{fig:posterAbstract}
	\end{figure}
	
	Previous theoretical work~\cite{Baumann2024} suggests that jump is the most effective mutation operator, which we confirm empirically. When given enough time, Jump-RLS/EA almost find nearly-optimal solutions, see Figure~1(b). We verify with statistical significance (using the Wilcoxon rank sum test~\cite{Wilcoxon1945}) that swaps are worse than exchange, which are in turn worse than jumps on the tested instances. The jump-operator also clearly outperforms all other state-of-the-art algorithms when given enough time. 
	%Nagamochi's algorithm overall shows the worst performance, which is somewhat surprising as it has the best performance guarantee. In practice, the achieved approximation ratios of the algorithms fall far below the theoretical threshold of $1.466 \cdot \opt$ (for $n=100$ vertices it is below $1.01 \cdot \opt$ for jumps on average), which might explain this phenomenon.
	
	While the jump-algorithms show the best performance, their running times are amongst the highest, see Figure~1. %From a computational perspective, the expected running time for computing the fitness change of a single jump is asymptotically the same as scanning for an \emph{acceptable} jump, that is, a jump that does not increase the number of crossings. 
	We improve the convergence-speed of Jump-RLS by not performing jumps at random, but by scanning for \emph{acceptable jumps} (i.e. not increasing the crossings number), which does not increase the expected running time asymptotically. We propose three different strategies to make a choice among the acceptable moves found by the algorithm: %In case no acceptable move is found, all algorithms retain the original solution. 
	Performing the first acceptable jump (\emph{JFIRLS}), scanning all jumps and selecting an acceptable one uniformly at random (\emph{JRIRLS}), and choosing the best jump (\emph{JSRLS}). We tested the three algorithms on the same datasets. We verified with statistical significance that the JFIRLS is worse than the other two variants, which show roughly the same performance, see Figure~1(c). %The convergence of the JRIRLS is a little slower, but the solutions tend to be a little better (but without statistical significance). This might be the case as the random moves of the JRIRLS enables a better exploration of plateaus. 
	The JRIRLS and the JSRLS converge up to 100 times faster than a normal Jump-RLS or Jump-EA on these instances, which coincides with a factor of $n$.% (in this case instead of roughly $10^5=n^2\sqrt n$ generations, we only needed $10^3=n\sqrt n$ until convergence).  

	%Concluding, we see that choosing the correct mutation operator is crucial for the design of EAs. Using the jump operator turns out to be far superior to exchanges and swaps, and by applying lightweight grey-box optimisation, we were able to design an algorithm that beats the state-of-the-art in terms of solution quality, while still admitting a decent running time.
	%From a more theoretical point of view, it is interesting that the JSRLS algorithm achieves a speed-up of a factor of roughly 100, which happens to coincide precisely with the size of~$\nY$ of our instances.  Could it be that the expected speed-up is indeed a linear factor~$\nY$?  Can this be proved rigorously?
	\bibliography{references}
	

\section*{Supplementary material}

# Evolutionary Algorithms for One-sided Bipartite Crossing Minimisation

Jakob Baumann • Ignaz Rutter • Dirk Sudholt

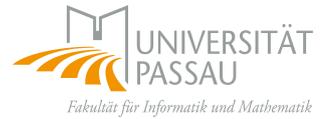

### Goals
- Understand EAs in the context of graph drawing by exploring a simple GD problem
- Complement theoretical findings in [1]
- Design specialised EA-based algorithm and compare it to the state-of-the-art

### ONE-SIDED BIPARTITE CROSSING MINIMIZATION (OBCM)

**Input:** Bipartite Graph $G = (X, Y, E)$
Permutation $\pi_X$ of $X$

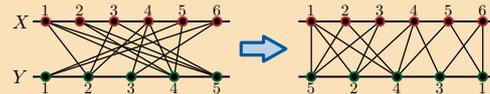

**Problem:** Find a permutation $\pi_Y$ minimising the number of edge crossings

### State-of-the-art algorithms
- **Median/Barycenter** $\Theta(n \log n + m)$: Place vertices at the average of their neighbors [2]
- **Nagamochi's algorithm** $\Theta(n \log n)$: Place vertices by some specific scheme; best theoretical bound [3]
- **Sifting** $\Theta(n^2 + nm)$: Sequentially place vertices at locally optimal positions [4]

### Common permutation-based mutation operators

**Swap:** Pick vertex (u.a.r.) and swap with right neighbor

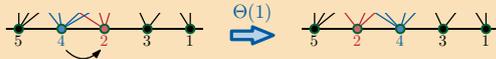 $\Theta(1)$

**Exchange:** Pick two vertices (u.a.r.) and exchange their positions

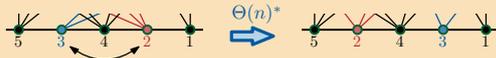 $\Theta(n)^*$

**Jump:** Pick a vertex (u.a.r.) and jump it to a new position (u.a.r.)

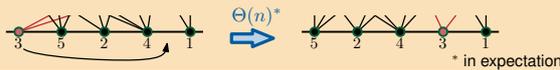 $\Theta(n)^*$

\* in expectation

### Evolutionary algorithms
- The **X-EA** is a typical (1+1)-EA with $X$ being one of the mutation operators swap, exchange, and jump

**Algorithm 1:** (1+1)-EA for permutation-based optimisation.
1. Choose permutation $\pi$ u.a.r.;
2. **while** *stopping criterion not met* **do**
3.   Choose $k$ by Poisson distribution with $\lambda = 1$;
4.   $\pi' \leftarrow$ apply mutation $X$ $(k+1)$-fold to $\pi$;
5.   **if** $c(\pi') \leq c(\pi)$ **then** $\pi \leftarrow \pi'$;

- **X-RLS** is a randomised local search variant, where $k = 1$ is constant

### Experimental Comparison
- Instances of size $100 + 100$, random edges (with density $0.02$ to $0.08$)
- Stopping criterion: no improvement for $n^{1.5}$ generations
- Preprocessing usually takes a significant amount of time
- Jump- and Exchange-EAs need most time but get best results
- Tested on different classes, results look alike
- Fitness over time plot reveals jump is by far the best! Taking roughly $n^{2.5}$ generations
- Performance differences are validated by Wilcoxon rank-sum test

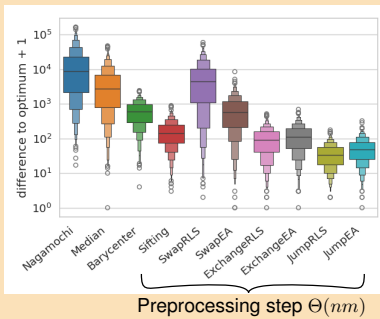

Preprocessing step $\Theta(nm)$

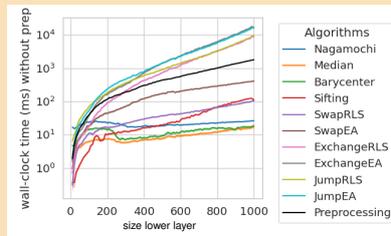

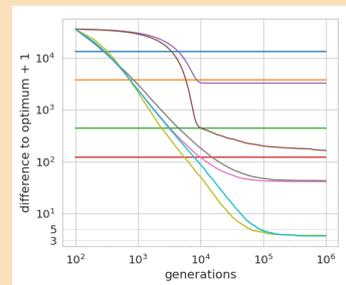

### Faster Convergence through Improved Algorithm Designs
- No overhead: instead of random jumps, scan for a fitness-improving jump
- Tested three variants: Select vertex u.a.r. then
  - **JFI-RLS**: Choose the first improving jump
  - **JRI-RLS**: Scan all jumps and choose an improving one u.a.r.
  - **JS-RLS**: scan all jumps and take the best (*sift*)
- Much better convergence rate for JRI- and JS-RLS, factor 100 (= $n$?)
- JFI-RLS statistically significantly worse, JS- and JRI-RLS roughly equal
- Very close approximation to optimum. JRI-RLS can better explore the plateau

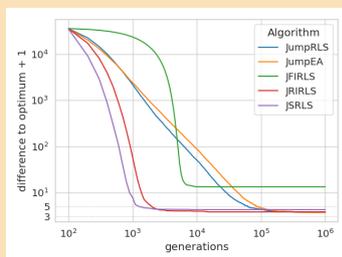

### Conclusion and Future Work
- Even simple "(1+1)"-type EA can beat best approx. algorithms (in reasonable time)
- Solutions become astonishingly close to the optimum
- Improvement with problem-specific knowledge yields very practical algorithm
- Can we prove an approximation ratio for jumps in poly time?
- Can the algorithm be improved using populations and crossover?
- Extend analysis to multiple layers, and see how EAs can actually be used to draw aesthetically pleasing graphs


	
\end{document}